\documentclass[11pt]{article}

\usepackage{acl}

\usepackage{times}
\usepackage{latexsym}
\usepackage[T1]{fontenc}
\usepackage[utf8]{inputenc}
\usepackage{microtype}
\usepackage{inconsolata}

\usepackage{amsmath,amssymb,amsfonts}
\usepackage{algorithm}
\usepackage{algorithmic}
\usepackage{graphicx}
\usepackage{booktabs}
\usepackage{pgfplots}
\pgfplotsset{compat=1.18}
\usepackage{subcaption}

\usepackage{tabularx}

\title{Latent Recurrent Transformer: Architecture Exploration, \\Training Strategies, and Scaling Behavior}

\author{
  \textbf{Zeyi Huang}\textsuperscript{1,2$^\star$},
  \textbf{Xuehai He}\textsuperscript{1$^\star$},
  \textbf{Liliang Ren}\textsuperscript{1},
  \textbf{Yiping Wang}\textsuperscript{3},
  \textbf{Baolin Peng}\textsuperscript{1},
  \textbf{Hao Cheng}\textsuperscript{1},
  \\
  \textbf{Shuohang Wang}\textsuperscript{1},
  \textbf{Pengcheng He}\textsuperscript{1},
  \textbf{Jianfeng Gao}\textsuperscript{1},
  \textbf{Yong Jae Lee}\textsuperscript{2$\dagger$},
  \textbf{Yelong Shen}\textsuperscript{1$\dagger$}
  \\[0.8em]
  \textsuperscript{1}Microsoft
  \hspace{0.6cm}
  \textsuperscript{2}University of Wisconsin--Madison
  \hspace{0.6cm}
  \textsuperscript{3}University of Washington
}

\begin{document}
\maketitle
\begingroup
\renewcommand{\thefootnote}{}
\footnotetext{
Work done during Zeyi's internship at Microsoft. Correspondence to \{zeyihuang,yongjaelee\}@cs.wisc.edu and 
\{xuehaihe,yeshe\}@microsoft.com. $^\star\dagger$ denote equal contribution and advising.
}
\endgroup

\begin{abstract}
We study \emph{Latent Recurrent Transformer} (LRT), a lightweight
augmentation of autoregressive transformers that reuses a high-level
source-layer hidden state from the previous token as recurrent memory for the
next token. Because this state is already computed during ordinary decoding,
LRT introduces a cross-token, cross-layer latent pathway while preserving the
standard attention mechanism, KV-cache interface, and one model forward per
generated token. To pretrain this recurrence without sequentially unrolling
the full sequence, we introduce \emph{interleaved parallel training}: one
full-sequence initialization forward constructs a shared buffer, followed by
sequential refinement of disjoint position subsets with parallel computation
within each subset. This provides every token with recurrent-memory-aware
supervision at approximately $2\times$ ideal token compute. 
Across 1.3B- and 2.1B-parameter nanochat-style backbones and a wide range of
training budgets, LRT improves both BPB and CORE under matched effective
compute. Additionally, LRT outperforms two-forward PonderLM-2 and matches a three-loop Transformer in BPB, while retaining one-forward-per-token decoding with $9\%$ latency overhead over the standard Transformer. Our code is publicly available \href{https://github.com/OoDBag/Latent_recurrent_transformer}{here}.

\end{abstract}

\section{Introduction}

\begin{figure}[t]
    \centering
    \includegraphics[width=0.48\textwidth]{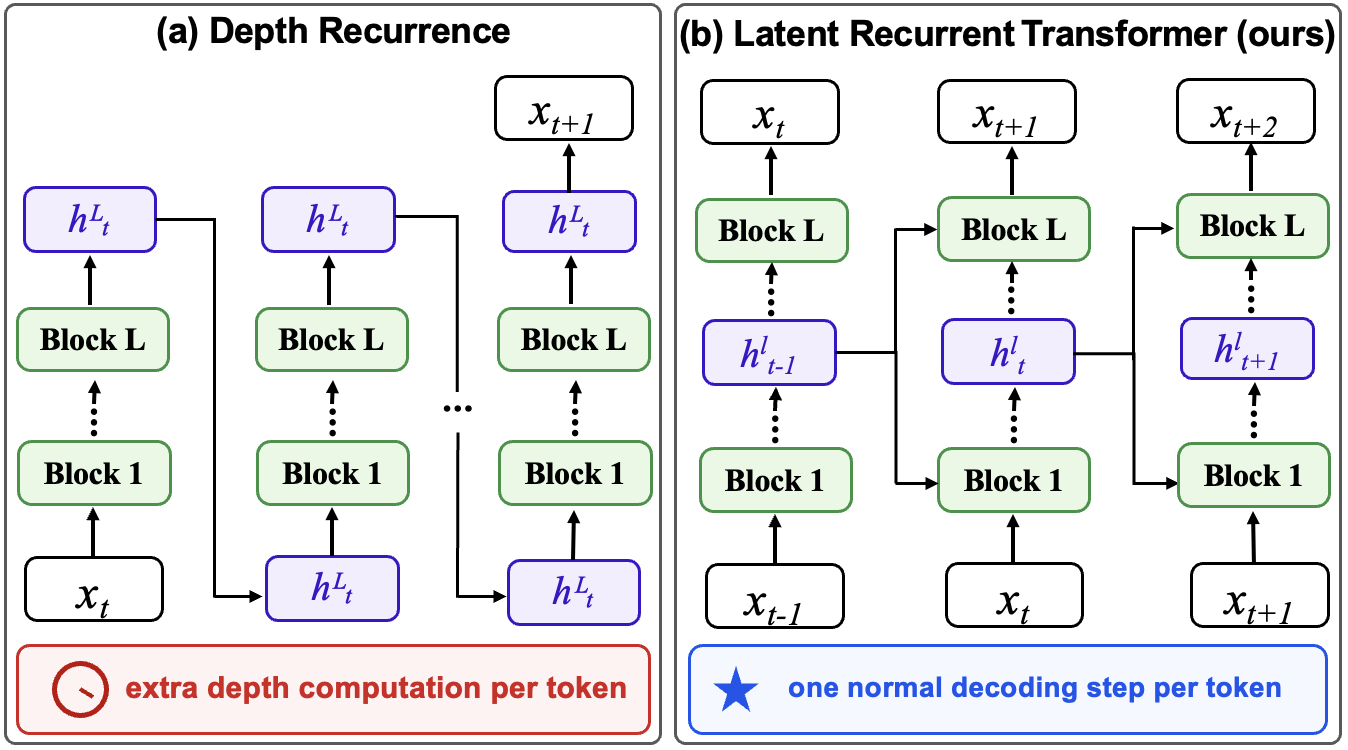}
    \caption{
    Overview of Latent Recurrent Transformer (LRT).
    \textbf{(a) Depth-recurrent methods} increase computation by repeatedly applying model blocks to the same token before emitting an output.
    \textbf{(b) LRT} reuses a high-level source-layer state as recurrent memory for the next token, creating a cross-token latent recurrence without adding extra autoregressive decoding steps. Each generated token still uses one normal transformer forward pass.
    }
    \label{fig:lrt_overview}
\end{figure}

Autoregressive transformers are the standard architecture for language modeling~\citep{vaswani2017attention,radford2019language,brown2020language}, but each generated token is still produced by a fixed-depth feedforward computation. A natural way to increase computation is to introduce recurrence. Existing approaches often add recurrence either in depth, by repeatedly applying blocks to the same token~\citep{dehghani2018universal,giannou2023looped}, or in time, by inserting pause or thinking tokens before emitting each real token~\citep{goyal2024think}. While these methods can enable iterative refinement, they also increase inference cost through extra block applications or additional decoding steps. 

We observe that autoregressive decoding already computes high-level recurrent signals for free: the hidden states of the previous token. Upper-layer states are trained toward next-token prediction and can provide useful latent context for processing the next position. This motivates a simple question: can we reuse an already-computed high-level representation from the previous tokens as recurrent memory, without adding extra decoding steps?

We study \emph{Latent Recurrent Transformer} (LRT), a lightweight augmentation of standard autoregressive transformers. At token position $t$, LRT reuses a source-layer hidden state from the previous position as recurrent memory, $\mathbf{m}_{t-1}=\mathbf{h}^{\ell_{\mathrm{src}}}_{t-1}$, where $\ell_{\mathrm{src}}$ is the source layer. It processes the usual token input with the standard KV cache, while injecting $\mathbf{m}_{t-1}$ into transformer layers through lightweight mechanisms such as \emph{KV Projection} and \emph{Residual Injection}. LRT preserves the decoder-only backbone, attention mechanism, feedforward layers, and KV-cache interface; the memory acts as an auxiliary latent pathway across adjacent autoregressive steps.

This latent pathway gives LRT a cross-layer route that standard causal attention lacks. In a KV-cached transformer, layer $\ell$ at position $t$ can attend to previous tokens only through cached states from the same layer $\ell$; early layers therefore cannot directly use higher-layer representations of previous tokens. By injecting the source-layer memory into the current token's layers, LRT lets early computation at position $t$ access higher-level information already computed for position $t-1$, while still using the standard KV cache and one normal forward pass per generated token.

The main challenge is pretraining. At inference time, LRT naturally forms a token-level recurrent chain, $\mathbf{m}_1 \rightarrow \mathbf{m}_2 \rightarrow \cdots \rightarrow \mathbf{m}_T$. Exactly reproducing this chain during training would require sequentially unrolling the transformer over the full sequence, destroying the parallelism that makes transformer pretraining efficient. Chunked training preserves parallelism within chunks, similar to segment- or block-level recurrent computation~\citep{dai2019transformer,hutchins2022block,sun2023retentive}, but recurrent memory is only propagated across chunk boundaries.

We therefore introduce \emph{interleaved parallel training}. An initialization forward first constructs a full-sequence KV and recurrent-state buffer. The model then refines disjoint interleaved subsets sequentially, with computation remaining parallel within each subset, and writes their updated states back to the shared buffer. This gives every position a recurrent-memory-aware refinement step at approximately $2\times$ ideal token compute. The same formulation also supports full-sequence refinement, in which all positions are updated synchronously over one or more refinement forwards, providing an additional training-compute axis without changing recurrent decoding.

Empirically, LRT shifts scaling curves toward lower bits per byte (BPB) and higher CORE under baseline-equivalent training compute across 1.3B- and 2.1B-parameter backbones. Furthermore, LRT outperforms two-forward PonderLM-2 and matches the BPB of a three-loop Transformer, while requiring only one inference forward per generated token. In our serving setup, this recurrent decoding introduces only $9\%$ latency overhead relative to the standard Transformer.

In summary, we make three contributions. First, we propose Latent Recurrent Transformer, a lightweight recurrent extension that passes a high-level source-layer state from one token to the next while preserving the standard KV-cache interface and one model forward per generated token. Second, we introduce interleaved parallel training and its full-sequence refinement configuration, enabling recurrent-memory-aware pretraining without exact token-by-token unrolling. Third, we show that LRT improves the compute quality trade-off over matched-depth Transformer baselines across model sizes and training budgets, while retaining one-forward-per-token decoding with low measured latency overhead.

%

\section{Latent Recurrent Transformer Architecture}
\label{sec:architecture}

Latent Recurrent Transformer (LRT) augments a standard autoregressive transformer with a lightweight recurrent memory across adjacent token positions. Let $L$ be the number of transformer layers, $d$ the hidden dimension, and $\mathbf{h}^{\ell}_t\in\mathbb{R}^d$ the hidden state at position $t$ after layer $\ell$. We choose a source layer $\ell_{\mathrm{src}}$ and define the recurrent memory as $\mathbf{m}_t=\mathbf{h}^{\ell_{\mathrm{src}}}_t$. This memory is already computed during ordinary autoregressive decoding and can be reused when processing the next token without adding extra decoding steps.

At position $t$, LRT processes the current token with the standard KV cache $\mathcal{C}_{<t}$ while additionally injecting $\mathbf{m}_{t-1}$ into the transformer:
\begin{equation}
    \mathbf{h}^L_t, \mathbf{z}_t, \mathbf{m}_t, \mathcal{C}_{\leq t}
    =
    f_\theta\!\left(x_t, \mathbf{m}_{t-1}, \mathcal{C}_{<t}\right),
\end{equation}
where $\mathbf{z}_t$ are the next-token logits, $\mathbf{m}_t=\mathbf{h}^{\ell_{\mathrm{src}}}_t$ is the updated recurrent memory, and $\mathcal{C}_{\leq t}$ is the updated KV cache. At the first position, $\mathbf{m}_{t-1}$ is initialized to zero.

LRT adds a cross-token, cross-layer pathway that is not available in a standard KV-cached transformer. Standard causal attention lets layer $\ell$ at position $t$ attend to cached states from previous positions at the same layer; LRT lets target layers at position $t$ access the source-layer memory from position $t-1$. Thus, even an intermediate source layer can provide high-level feedback to earlier layers of the next token. The standard attention mechanism and KV-cache interface are preserved: LRT adds no memory tokens and does not change the cache shape, but only changes how current-token layer inputs or key-value vectors are formed.

Our default LRT combines two lightweight injection mechanisms: \emph{KV Projection}, which injects memory through the attention key-value pathway, and \emph{Residual Injection}, which adds memory to the residual stream. Architecture ablations in Section~\ref{sec:ablations} show that the two pathways are complementary.

\begin{figure*}[t!]
    \centering
    \includegraphics[width=0.99\textwidth]{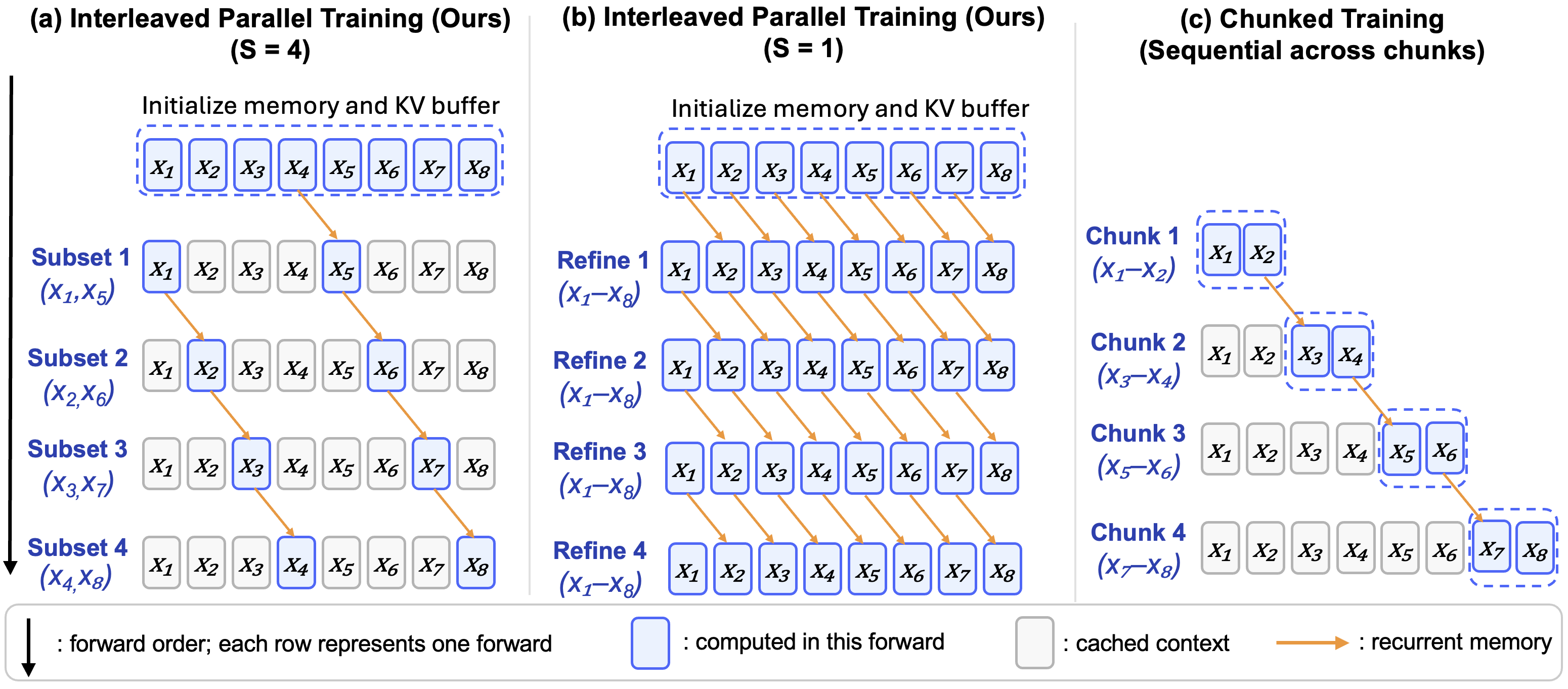}
\caption{
Training approximations for LRT. \textbf{(a) Interleaved parallel training} first performs a full initialization pass to populate a sequence-level KV and recurrent-state buffer, then refines $S$ disjoint interleaved subsets of positions. Updated states are written back to the buffer, allowing later subsets to consume recurrent memory refined by earlier subsets. The $S$ subset forwards jointly recompute the sequence once. \textbf{(b) With $S=1$}, the sole subset contains the entire sequence. We refer to each subsequent update as a \emph{full-sequence refinement} and use $K$ to denote the number of refinement forwards following initialization. All positions are updated synchronously from the buffer produced by the previous forward; the figure illustrates $K=4$. \textbf{(c) Chunked training} processes contiguous chunks sequentially. Computation remains parallel within each chunk, but recurrent memory is propagated only across chunk boundaries, giving a coarser approximation to token-level recurrence. Each row represents one forward pass. Blue boxes indicate recomputed positions, gray boxes indicate cached context, and orange arrows indicate recurrent memory or updated hidden states.
}
\label{fig:training_strategies}
\end{figure*}

\subsection{KV Projection}
\label{sec:kv_projection}

KV Projection injects recurrent memory through the attention key-value pathway, giving the previous token's source-layer representation a direct route into attention. For layer $\ell$, let $\mathbf{x}^{(\ell)}_t \in \mathbb{R}^d$ be the layer input, and let $\tilde{\mathbf{k}}^{(\ell)}_{\mathrm{local}}$ and $\mathbf{v}^{(\ell)}_{\mathrm{local}}$ be the locally computed raw key and value before positional encoding. We project the recurrent memory into the layer's key and value spaces:
\begin{equation}
    \tilde{\mathbf{k}}^{(\ell)}_{\mathrm{rec}}
    =
    W_{k,\mathrm{rec}} \mathbf{m}_{t-1},
    \quad
    \mathbf{v}^{(\ell)}_{\mathrm{rec}}
    =
    W_{v,\mathrm{rec}} \mathbf{m}_{t-1}.
\end{equation}
Here $n_{\mathrm{kv}}$ is the number of key-value heads and $d_{\mathrm{head}}$ is the head dimension, so 
$W_{k,\mathrm{rec}}, W_{v,\mathrm{rec}} \in \mathbb{R}^{(n_{\mathrm{kv}} d_{\mathrm{head}})\times d}$; 
the outputs are reshaped into $n_{\mathrm{kv}}$ heads of dimension $d_{\mathrm{head}}$.

We use input-dependent per-head gates to combine local and recurrent pathways:
\begin{equation}
    \mathbf{g}^{(\ell)}_{\mathrm{local},t},
    \mathbf{g}^{(\ell)}_{\mathrm{rec},t}
    =
    2 \cdot \sigma\!\left(
    W^{(\ell)}_g \mathbf{x}^{(\ell)}_t
    \right),
\end{equation}
where $\sigma$ is sigmoid and $W_g^{(\ell)} \in \mathbb{R}^{2n_{\mathrm{kv}} \times d}$. The output is split into local and recurrent gates, broadcast over the head dimension, and initialized to the neutral value $1$ by zero-initializing $W_g^{(\ell)}$.

The local and recurrent keys and values are combined as
\begin{equation}
\begin{aligned}
\tilde{\mathbf{k}}^{(\ell)}_t
&=
\mathbf{g}^{(\ell)}_{\mathrm{local},t} \odot \tilde{\mathbf{k}}^{(\ell)}_{\mathrm{local}}  +
\mathbf{g}^{(\ell)}_{\mathrm{rec},t} \odot \tilde{\mathbf{k}}^{(\ell)}_{\mathrm{rec}}, \\
\mathbf{v}^{(\ell)}_t
&=
\mathbf{g}^{(\ell)}_{\mathrm{local},t} \odot \mathbf{v}^{(\ell)}_{\mathrm{local}}  +
\mathbf{g}^{(\ell)}_{\mathrm{rec},t} \odot \mathbf{v}^{(\ell)}_{\mathrm{rec}} .
\end{aligned}
\end{equation}
The combined raw key follows the same QK normalization and RoPE pipeline as the standard key,
$\mathbf{k}^{(\ell)}_t=\mathrm{RoPE}_t(\mathrm{QKNorm}(\tilde{\mathbf{k}}^{(\ell)}_t))$,
so the recurrent key inherits the position-$t$ encoding. The resulting keys and values have the same shape as standard attention KV tensors and use the usual KV-cache interface. We use additive composition; replacing local KV features with recurrent projections underperforms in Appendix~\ref{app:architecture_ablation}.

\subsection{Residual Injection}
\label{sec:residual_injection}

Residual Injection exposes each transformer block to recurrent memory through the residual stream. For block input $\mathbf{x}^{(\ell)}_t$, we form
\begin{equation}
    \bar{\mathbf{x}}^{(\ell)}_t
    =
    \alpha_\ell \mathbf{x}^{(\ell)}_t
    +
    \gamma_\ell \mathbf{m}_{t-1},
\end{equation}
where $\alpha_\ell$ is the same learnable residual scale used in the baseline block and is initialized to $1.0$, while $\gamma_\ell$ is the LRT memory scale initialized to $0.1$. For the pre-norm backbone, $\bar{\mathbf{x}}^{(\ell)}_t$ is used as the block input before RMSNorm, attention, and MLP computations.

Residual Injection adds minimal overhead and gives every block direct access to the recurrent memory. Unlike KV Projection, it mixes memory into the general residual stream rather than giving it a dedicated attention pathway. We combine both mechanisms in the default LRT, which performs best in our architecture ablations.


\section{Training Latent Recurrent Transformer}
\label{sec:training}

At inference time, LRT naturally forms a token-level recurrent chain: after
processing token $t$, the model stores
$\mathbf{m}_t=\mathbf{h}^{\ell_{\mathrm{src}}}_t$ and reuses it for token
$t+1$. Exact training would require sequentially unrolling this chain over the
full sequence, since each token depends on the recurrent state produced by its
predecessor. This resembles backpropagation through
time~\citep{werbos1990backpropagation} and would eliminate the parallelism that
makes Transformer pretraining efficient.

We therefore seek a parallel approximation that gives each token a recurrent-memory-aware training signal. We introduce \emph{interleaved parallel training}, which first builds a full-sequence buffer and then refines token subsets using recurrent states stored in the buffer. Figure~\ref{fig:training_strategies} illustrates configurations with multiple interleaved subsets and with a single full-sequence subset. We additionally compare with sequential chunk training, which preserves parallelism within contiguous chunks but propagates recurrent memory only across chunk boundaries.

\subsection{Interleaved Parallel Training}
\label{sec:interleaved_training}

Interleaved parallel training uses one full-sequence initialization pass
followed by sparse refinement over $S$ disjoint subsets. We partition positions
$\{1,\dots,T\}$ into subsets
$\mathcal{I}_1,\dots,\mathcal{I}_S$. The initialization pass processes all
positions in parallel and builds a buffer $\mathcal{B}^{(0)}$ containing
per-layer keys, values, and source-layer recurrent states. Then, for
$s=1,\dots,S$, we recompute only positions in $\mathcal{I}_s$ using the buffer
$\mathcal{B}^{(s-1)}$ and write the refined states back to form
$\mathcal{B}^{(s)}$.

The write-back step lets later subsets read recurrent memory refined by earlier
subsets, forming a sparse recurrent chain while keeping each subset forward
parallel. Across all refinement steps, every position is recomputed once with
recurrent memory from the shared buffer. Thus, in ideal token-compute terms,
training costs approximately $2\times$ a standard Transformer update: one
initialization pass plus one effective refinement pass.

We use $S=2$ by default with a strided partition, where subset
$\mathcal{I}_s$ contains positions $s,s+S,s+2S,\ldots$. The training objective
averages the initialization and per-subset refinement losses,
$\mathcal{L}=(\mathcal{L}_{\mathrm{init}}+
\sum_{s=1}^{S}\mathcal{L}_{\mathcal{I}_s})/(S+1)$, where
$\mathcal{L}_{\mathrm{init}}$ is the full-sequence cross-entropy and
$\mathcal{L}_{\mathcal{I}_s}$ is the cross-entropy over positions in
$\mathcal{I}_s$. The initialization loss preserves a standard training signal,
while the refinement losses train the model to use recurrent memory from the
shared buffer. We backpropagate jointly through the initialization and
refinement forwards. 

\paragraph{Full-sequence refinement for $S=1$.}
The same formulation can be used with $S=1$, where the sole subset contains
the entire sequence, $\mathcal{I}_1=\{1,\ldots,T\}$. Each refinement forward
therefore recomputes all positions in parallel. Unlike the default interleaved
schedule, all positions are updated synchronously, so states produced by one
refinement forward become available only in the next forward. 

To allow repeated full-sequence refinement, we use $K$ to denote the number of
refinement forwards following initialization. Starting from the initialization
buffer $\mathcal{B}^{(0)}$, refinement forward $k$ recomputes the entire
sequence using $\mathcal{B}^{(k-1)}$ and writes the resulting states to
$\mathcal{B}^{(k)}$. By default, we compute a loss on the initialization and
every refinement forward and average them uniformly,
$\mathcal{L}=(\mathcal{L}_{\mathrm{init}}+
\sum_{k=1}^{K}\mathcal{L}_{\mathrm{full}}^{(k)})/(K+1)$. We backpropagate
jointly through all $K+1$ forwards without detaching the intermediate buffers.

With $K=1$, full-sequence refinement has the same ideal $2\times$ token compute as the default interleaved schedule; in general, its cost is approximately $(K+1)\times$ that of a standard Transformer update. Comparisons with interleaved and exact sequential training, together with the ablation on $K$, are provided in Appendix~\ref{app:parallel_training_analysis}.

\subsection{Comparison to Chunked Training}
\label{sec:chunked_training}

We also consider chunked training as a lower-token-compute approximation inspired by segment-level and blockwise recurrent models~\citep{dai2019transformer,hutchins2022block,sun2023retentive}. As shown in Figure~\ref{fig:training_strategies}(c), the sequence is split into contiguous chunks that are processed sequentially, while computation remains parallel within each chunk.

Chunked training is close to standard transformer training in ideal token compute because each token is processed once. However, it gives a coarse approximation to token-level recurrence: memory is propagated only across chunk boundaries, so tokens inside the same chunk do not receive recurrent memory from their immediate predecessors. In contrast, interleaved parallel training refines disjoint token subsets so that every token receives a recurrent-memory-aware refinement loss. Moreover, chunked training can be slower in wall-clock time than its ideal token-compute estimate suggests, since each sequence requires multiple sequential chunk forwards. Our experiments show that it remains weaker than interleaved parallel training.

\begin{figure*}[t]
    \centering
    \begin{subfigure}[t]{0.49\linewidth}
        \centering
        \includegraphics[width=\linewidth]{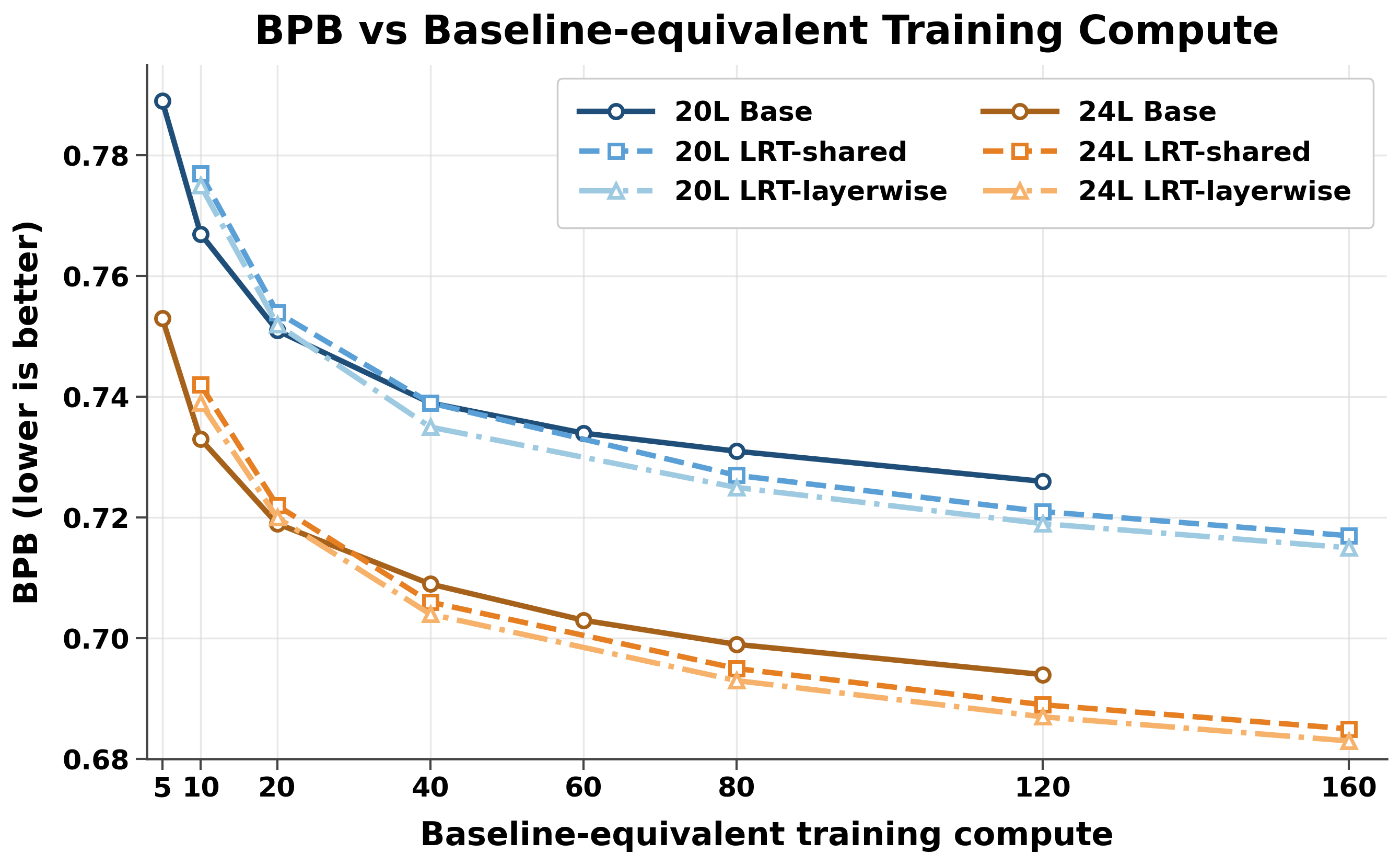}
        \caption{Language modeling.}
        \label{fig:scaling_bpb}
    \end{subfigure}
    \hfill
    \begin{subfigure}[t]{0.49\linewidth}
        \centering
        \includegraphics[width=\linewidth]{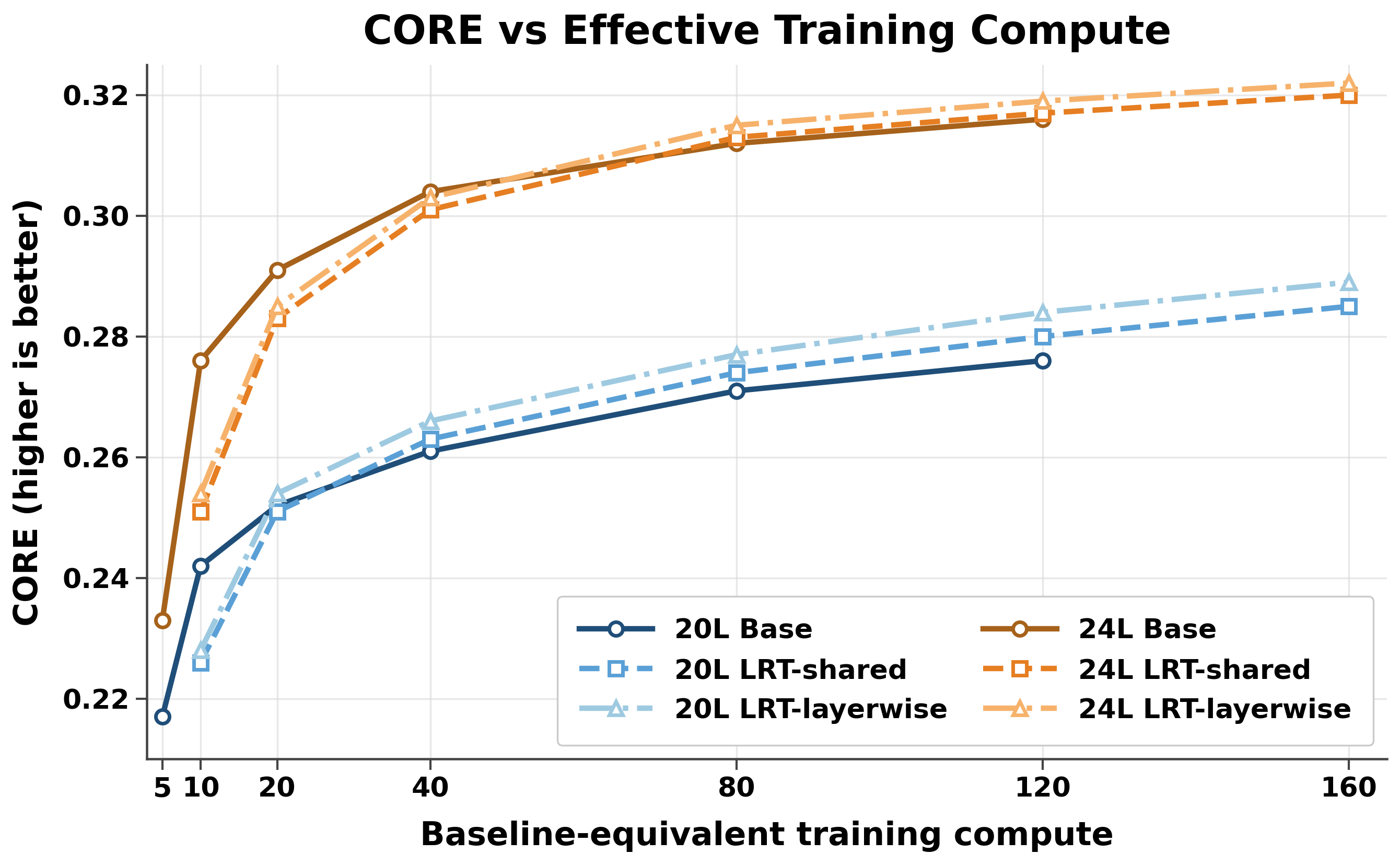}
        \caption{In-context evaluation.}
        \label{fig:scaling_core}
    \end{subfigure}

\caption{
Scaling behavior of Latent Recurrent Transformers under baseline-equivalent training compute.
\textbf{Left:} BPB versus training compute, where lower is better.
\textbf{Right:} CORE versus training compute, where higher is better.
We plot methods by baseline-equivalent training compute: $R$ denotes the tokens-per-parameter training budget. A standard transformer trained at ratio $R$ is plotted at $R$, while an LRT trained with interleaved parallel training at ratio $R$ is plotted at $2R$ because it uses one initialization pass and one effective refinement pass.
\textbf{LRT-shared} is our default lightweight variant with recurrent projections shared across layers, adding $0.3\%$ parameters.
\textbf{LRT-layerwise} uses separate recurrent projections per layer, adding $4.8\%$ parameters for 20L and $5.4\%$ for 24L.
Across both depths, LRT shifts the scaling curves toward lower BPB and higher CORE, with the layerwise variant providing additional gains at higher parameter cost.
}
\label{fig:scaling_effective_compute}
\end{figure*}

\section{Experiments}
\label{sec:experiments}

We build on nanochat\footnote{\url{https://github.com/karpathy/nanochat}}~\citep{karpathy2025nanochat}, a reproducible modern GPT-style pretraining stack. To isolate the effect of LRT, we keep the baseline implementation, data pipeline, tokenizer, optimizer, batching scheme, and evaluation protocol aligned with nanochat. LRT modifies only the transformer through its recurrent memory pathway and interleaved parallel training.

\paragraph{Implementation details.}
We use the nanochat-style GPT backbone, including parameter-free RMSNorm~\citep{zhang2019root}, RoPE~\citep{su2024roformer}, squared-ReLU MLPs, multi-head self-attention with sliding-window attention, value embeddings, and logit soft-capping. Unless otherwise specified, LRT uses combined KV Projection and Residual Injection, dual gating, zero memory initialization, and interleaved parallel training with $S=2$. We report two variants: \emph{LRT-shared}, the default lightweight model with recurrent KV projections shared across layers, and \emph{LRT-layerwise}, a higher-capacity model with separate recurrent projections per layer. LRT-shared adds $0.3\%$ parameters; LRT-layerwise adds $4.8\%$ for 20L and $5.4\%$ for 24L. Full implementation details are in Appendix~\ref{app:implementation}.

\paragraph{Training data and optimization.}
Following nanochat~\citep{karpathy2025nanochat}, we pretrain on FineWeb-Edu 100BT~\citep{lozhkovfineweb} using the pre-shuffled nanochat release and nanochat BPE tokenizer. We train with MuonAdamW~\citep{jordan2024muon}, global batch $2^{19}$ tokens, and sequence length $2048$, holding out the final shard for validation. Optimizer hyperparameters, precision, hardware, schedule, and FLOP accounting are in Appendix~\ref{app:implementation}.

\paragraph{Metrics.}
We report bits per byte (BPB; lower is better), a tokenization-independent language-modeling metric used in modern data-scaling evaluations~\citep{li2024datacomp,karpathy2025nanochat}. BPB normalizes total cross-entropy by target bytes, $\mathrm{BPB}=\sum_i \ell_i / (\ln 2 \cdot \sum_i b_i)$, where $\ell_i$ is token-level cross-entropy and $b_i$ is token byte length. Although BPB differences may appear small, BPB is already a log-loss quantity normalized per byte; we therefore report it directly and interpret improvements through consistent reductions and matched-compute comparisons. Appendix~\ref{app:baseline_ablations} further shows that generic architectural additions such as gated attention~\citep{qiu2026gated} and layer scaling~\citep{touvron2021going} yield no gains on the same strong baseline.

We also report CORE using the nanochat implementation of the DCLM evaluation suite~\citep{li2024datacomp,karpathy2025nanochat}. CORE averages centered scores over 22 few-shot in-context learning tasks covering language understanding, world knowledge, commonsense reasoning, symbolic problem solving, and reading comprehension. Following DCLM, each task score is centered relative to its random baseline, $\mathrm{centered}=(a-r)/(1-r)$, where $a$ is task accuracy and $r$ is the random baseline. CORE is the mean centered score across tasks. The full task list is in Appendix~\ref{app:core_details}. Unless otherwise specified, BPB and CORE results are means over three
independent runs; standard deviations are reported in
Appendix~\ref{app:statistical_reliability}.

\subsection{Setup}

We train two nanochat-style GPT backbones: \emph{20L} with $L=20$, $d=1280$, and approximately $1.3$B parameters, and \emph{24L} with $L=24$, $d=1536$, and approximately $2.1$B parameters. Baselines use the same backbone and training setup but remove the LRT recurrent memory modules.

We vary the tokens-per-parameter budget $R$: a model with $N$ trainable parameters trained at ratio $R$ sees approximately $R\!\times\!N$ training tokens. 
For example, a $1.3$B-parameter 20L model trained at $R=10$ sees about $13$B tokens. Baselines are trained up to $R=120$, and LRT variants up to $R=80$. Since interleaved parallel training uses one initialization pass plus one effective refinement pass, an LRT trained at ratio $R$ costs approximately the same as a baseline trained at ratio $2R$. We therefore plot baselines at baseline-equivalent compute $R$ and LRT at $2R$. Because FineWeb-Edu 100BT~\citep{lozhkovfineweb} is larger than our training budgets, this comparison is conservative for LRT in unique-token exposure: at the same baseline-equivalent compute, a baseline trained at ratio $2R$ sees roughly twice as many non-repeated tokens as an LRT trained at ratio $R$.

\subsection{Scaling Results}
\label{sec:scaling_results}

Figure~\ref{fig:scaling_effective_compute} plots BPB and CORE against baseline-equivalent training compute. This visualization aligns methods by approximate training cost rather than raw tokens-per-parameter ratio: the baseline has cost $1\times$, while LRT variants have cost approximately $2\times$ due to interleaved parallel training. Full numeric results are provided in Appendix~\ref{app:scaling_tables}; here we focus on the scaling trends.

Across both model depths, LRT shifts the scaling curve in the favorable direction: lower BPB and higher CORE at comparable baseline-equivalent compute. As discussed in the setup, this comparison is conservative for LRT in unique-token exposure, so the gains are unlikely to be explained by seeing more data. Instead, they suggest that the recurrent memory pathway helps use each training example and unit of compute more effectively.

On BPB, both LRT-shared and LRT-layerwise consistently improve over the matched-depth baseline across the scaling range. The curves show that the gain is not confined to a single training budget: LRT maintains a lower BPB trajectory as compute increases, even as all methods enter the slower-improvement regime at larger budgets. For example, on the 24L backbone, LRT-shared at baseline-equivalent compute $80$ reaches 0.695 BPB, improving over the 24L baseline at the same compute, which obtains 0.699. The layerwise variant improves further to 0.693.

CORE shows a similar trend. LRT improves aggregate in-context evaluation across most effective compute budgets, indicating that the recurrent latent pathway benefits not only token-level language modeling but also downstream in-context evaluation. For example, on the 20L backbone at baseline-equivalent compute $80$, the baseline obtains 0.271 CORE, while LRT-shared and LRT-layerwise obtain 0.274 and 0.277, respectively.

The comparison between LRT-shared and LRT-layerwise highlights the parameter--quality trade-off. LRT-shared is our default because it captures most of the benefit while adding only $0.3\%$ parameters. LRT-layerwise gives additional improvements by using separate recurrent projections per layer, but increases parameter overhead to $4.8\%$ for 20L and $5.4\%$ for 24L. We therefore treat LRT-layerwise as a higher-capacity variant and use LRT-shared as the default lightweight model.

\subsection{Comparison with Depth Recurrent and Looped Baselines}
\label{sec:iterative_baselines}

We compare LRT-layerwise with approaches that allocate additional model forwards to
each generated token. Loop Transformers repeatedly apply the model before
producing the next token, while PonderLM-2~\citep{zeng2025ponderlm} generates
an intermediate continuous thought before token prediction. Both therefore
increase inference computation with the number of loops or latent thoughts.
In contrast, LRT propagates recurrent memory across tokens while retaining one
model forward per generated token.

As shown in Table~\ref{tab:compare_other}, the default interleaved LRT reaches
0.752 BPB with one inference forward per token, slightly outperforming the
two-forward Loop Transformer and PonderLM-2 variants. LRT trained with
full-sequence refinement at $K=3$ further reaches 0.749 BPB, matching the
three-forward Loop Transformer while still using one inference forward per
token. Here, $K=3$ introduces additional refinement only during training:
the resulting model uses the same one-forward recurrent decoding procedure
as the default LRT. This configuration is included to show that additional
training-time refinement can improve LRT without increasing its inference
depth.

\begin{table}[t]
\centering
\small
\setlength{\tabcolsep}{4pt}
\begin{tabular}{@{}lcc@{}}
\toprule
\textbf{Method} & \textbf{\# Fwd./token} & \textbf{BPB} $\downarrow$ \\
\midrule
Transformer baseline
& $1$ & 0.767 \\
Loop Transformer (2 loops)
& $2$ & 0.753 \\
Loop Transformer (3 loops)
& $3$ & \textbf{0.749} \\
PonderLM-2 (1 thought)
& $2$ & 0.753 \\
\midrule
LRT, interleaved ($S=2$)
& $1$ & 0.752 \\
LRT, full-seq.\ ($K=3$)
& $1$ & \textbf{0.749} \\
\bottomrule
\end{tabular}
\caption{
Comparison with Depth Recurrent and Looped Baselines under the same 20-layer
pretraining setup. Both LRT rows use LRT-layerwise. ``\# Fwd./token'' denotes the number of complete model
forwards required to generate each new token during inference. Loop Transformer and PonderLM-2
improve BPB by increasing inference computation, whereas both LRT variants
retain one forward per token. Full-sequence refinement with $K=3$ uses
additional refinement forwards only during training. Lower BPB is better.
}
\label{tab:compare_other}
\end{table}

\subsection{Parallel Prefill and Recurrent Decoding}
\label{sec:parallel_prefill}

LRT supports two parallel strategies for prompt prefill. One-pass prefill
processes the prompt in a single parallel forward, whereas interleaved prefill
applies the same subset-refinement schedule used during training. In both
cases, the source-layer state at the final prompt position initializes
recurrent decoding. The prefill schedule is not tied to the model architecture
and can be selected according to the desired latency--quality trade-off.

In our evaluation, the choice of prefill strategy has little effect on
continuation quality. When BPB is measured over the 1,024 tokens following a
1,024-token prompt, both strategies remain within $0.001$ BPB of exact
sequential prefill: interleaved prefill differs by $+0.00002$, while one-pass
prefill differs by $+0.00082$. Thus, although interleaved prefill more closely
reproduces exact recurrence, both strategies provide nearly identical
continuation quality in this setting. One-pass prefill matches the Transformer
baseline in throughput, latency, and peak memory, whereas interleaved prefill
requires additional refinement forwards.

After prefill, LRT decodes autoregressively by reusing the source-layer state produced for each token as recurrent memory for the next token. Each generated token requires one model forward,
the standard KV-cache shape is preserved, and only one additional
$d$-dimensional recurrent state is stored per sequence. In our measurements,
LRT achieves $0.92\times$ decoding throughput and $1.09\times$ latency
relative to the Transformer baseline, with no measured increase in peak
memory. The complete evaluation setup and efficiency measurements are
provided in Appendix~\ref{app:efficiency}.

\subsection{Ablation Studies}
\label{sec:ablations}

We conduct ablations to understand three design choices in LRT: 
(1) which source layer should provide recurrent memory,
(2) which temporal state should serve as memory, and
(3) how this memory should be injected into the transformer.
Unless otherwise specified, ablations are conducted on the 20L model at ratio 10.

\paragraph{Source layer.}
We first ablate which hidden layer from the previous token should provide recurrent memory. Table~\ref{tab:source_layer_ablation} shows that an upper-middle layer, layer 12 in the 20L model, performs best. This suggests that useful recurrent memory should be high-level but not overly specialized for logits.

\begin{table}[t]
\centering
\small
\begin{tabular}{lc}
\toprule
\textbf{Source layer $\ell_{\mathrm{src}}$} & \textbf{BPB} $\downarrow$ \\
\midrule
Baseline / no recurrent memory & 0.767 \\
Layer 8 & 0.757 \\
Layer 12 & \textbf{0.754} \\
Layer 16 & 0.756 \\
Layer 20 (final layer) & 0.755 \\
Learned Avg. above 4 layers & \textbf{0.754} \\
\bottomrule
\end{tabular}
\caption{
Source-layer ablation on the 20L model at $R=10$. An upper-middle source layer performs best, while the final layer is not optimal. Lower BPB is better.
}
\label{tab:source_layer_ablation}
\end{table}

This result also clarifies why an intermediate source layer is not redundant with standard causal attention. Standard attention gives layer $\ell$ at position $t$ access to previous positions' cached states from the same layer, whereas LRT allows early target layers to receive a higher-level source representation $\mathbf{h}^{\ell_{\mathrm{src}}}_{t-1}$ from the previous token. Thus, an intermediate layer can provide useful feedback to earlier layers at the next position. For the 24L model, we follow the same relative-depth choice and use layer 14 as source layer.

\paragraph{Temporal memory source.}
We next fix the source layer to $\ell_{\mathrm{src}}=12$ and ablate which
past temporal state should be used as recurrent memory. Our default choice
uses the immediately preceding source-layer state,
$\mathbf{m}_{t-1}=\mathbf{h}^{\ell_{\mathrm{src}}}_{t-1}$, which is already
available during ordinary autoregressive decoding. Table~\ref{tab:memory_source_ablation}
compares this choice with older past states and a learned average of the four
most recent states.

\begin{table}[t]
\centering
\small
\setlength{\tabcolsep}{4pt}
\begin{tabularx}{\columnwidth}{Xc}
\toprule
\textbf{Memory source} & \textbf{BPB} $\downarrow$ \\
\midrule
Baseline / no recurrent memory & 0.767 \\
$\mathbf{m}_{t-4}$ & 0.760 \\
$\mathbf{m}_{t-3}$ & 0.760 \\
$\mathbf{m}_{t-2}$ & 0.757 \\
$\mathbf{m}_{t-1}$ & \textbf{0.754} \\
Learned avg.\ of previous four states & \textbf{0.754} \\
\bottomrule
\end{tabularx}
\caption{
Temporal memory-source ablation on the 20L model at $R=10$. The immediately preceding state $\mathbf{m}_{t-1}$ is the strongest single source and matches a learned average of four recent states. Lower BPB is better.
}
\label{tab:memory_source_ablation}
\end{table}

Among the evaluated memory sources, $\mathbf{m}_{t-1}$ is the strongest
single state. Performance degrades as the temporal distance increases,
indicating that the immediately preceding source-layer representation provides
the most useful recurrent signal. A learned average over multiple recent
states does not improve over $\mathbf{m}_{t-1}$ alone, suggesting that
explicitly aggregating a longer temporal history is unnecessary in this
setting. We therefore use $\mathbf{m}_{t-1}$ as the default memory source.

\paragraph{Memory injection.}
We next ablate how recurrent memory enters the transformer. Residual Injection adds memory to the residual stream, while KV Projection maps it into the attention key-value pathway. Both improve over the baseline: KV Projection gives the previous token's source representation a targeted attention route, while Residual Injection exposes it to the whole block computation. Their combination performs best, suggesting complementary attention-level and residual-stream pathways. Detailed KV variants are provided in Appendix~\ref{app:architecture_ablation}; shared versus layerwise projections are compared in the full scaling tables.

\begin{table}[t]
\centering
\small
\begin{tabular}{lc}
\toprule
\textbf{Architecture} & \textbf{BPB} $\downarrow$ \\
\midrule
Baseline / no recurrent memory & 0.767 \\
Residual Injection & 0.756 \\
KV Projection & 0.755 \\
KV Projection + Residual Injection & \textbf{0.754} \\
\bottomrule
\end{tabular}
\caption{
Memory-injection ablation on the 20L model at $R=10$. Combining KV Projection and Residual Injection performs best. Lower BPB is better.
}
\label{tab:memory_injection_ablation}
\end{table}

\subsection{SFT and Reinforcement Learning}
\label{sec:sft_rl}

Supervised fine-tuning requires no modification and follows the same parallel training procedure as pretraining. Reinforcement learning introduces a mismatch because recurrent states are produced autoregressively during rollout but policy probabilities are recomputed using parallel refinement forwards. We reduce this mismatch by caching the rollout recurrent states and using them, without gradients, to initialize the recurrent-state buffer during policy recomputation. On GSM8K with GRPO, this initialization improves LRT from $12\%/41\%$ to $14\%/50\%$ pass@1/pass@16, compared with $7\%/38\%$ for the Transformer baseline. 

\section{Related Work}
\label{sec:related}

\paragraph{Depth recurrence and iterative computation.}
A common way to increase per-token computation is to add extra work before emitting each token. Depth-recurrent methods, including Universal Transformers~\citep{dehghani2018universal}, looped transformers~\citep{giannou2023looped}, and Huginn~\citep{geiping2026scaling}, repeatedly apply layers or blocks to the same token. Other methods introduce auxiliary discrete or continuous computation~\citep{pfau2024let,herel2024thinking,zelikman2024quiet,hao2024training}. PonderLM-2~\citep{zeng2025ponderlm} generates latent thoughts between ordinary tokens and feeds them back through the model, whereas PCCoT~\citep{wu2025parallel} uses Jacobi iteration to refine inserted continuous thought tokens in parallel. These methods perform additional computation on the current token or auxiliary latent positions at inference. In contrast, LRT reuses a high-level representation already computed for the previous token. Its refinement forwards are used only to train the recurrent pathway in parallel and are not required during autoregressive decoding, which uses one normal Transformer forward per generated token.

\paragraph{Memory and chunked recurrent transformers.}
Many transformer variants add recurrence or memory across time, segments, or blocks. Feedback Transformers~\citep{fan2021feedback} expose high-level past representations to future computation; Transformer-XL~\citep{dai2019transformer} and Compressive Transformers~\citep{rae2019compressive} reuse or compress segment-level states; and memory-augmented models introduce memory tokens, block-level states, external retrieval, landmark anchors, or compressive working memory~\citep{bulatov2022recurrent,hutchins2022block,wu2022memorizing,hwang2024transformerfam,mohtashami2023landmark,munkhdalai2024leave}. Other recent approaches update test-time memory online~\citep{sun2024learning,behrouz2026titans}, while chunked or linear recurrent architectures balance recurrence and parallelism through blockwise computation~\citep{pilault2023block,sun2023retentive}. These methods often target longer context, persistent state, or long-sequence efficiency. LRT instead uses the immediately preceding token's source-layer hidden state, $\mathbf{m}_{t-1}=\mathbf{h}^{\ell_{\mathrm{src}}}_{t-1}$, as a lightweight token-level latent memory inside a standard KV-cached autoregressive transformer. Our chunked baseline propagates memory only across chunk boundaries, while interleaved parallel training refines disjoint token subsets and writes states back to a shared buffer, giving every token a recurrent-memory-aware refinement step while retaining substantial parallelism.

\paragraph{Recurrent sequence models.}
Another line of work revisits recurrence as an alternative to attention. Linear Transformers~\citep{katharopoulos2020transformers} express attention as a linear recurrence, while state-space and convolutional models such as S4~\citep{gu2021efficiently}, H3~\citep{fu2022hungry}, S5~\citep{smith2022simplified}, Mamba~\citep{gu2023mamba}, and Hyena~\citep{poli2023hyena} use long-range operators with efficient recurrent forms. Recent gated recurrent architectures, including RWKV~\citep{peng2023rwkv}, Griffin/Hawk~\citep{de2024griffin}, and xLSTM~\citep{beck2024xlstm}, further improve recurrent language modeling, and hybrid models such as Jamba~\citep{lieber2024jamba} interleave attention and recurrence. LRT is complementary: it keeps the standard transformer backbone and KV cache, but adds a small recurrent channel that passes high-level latent features from one token to the next.

\section{Conclusion and Future Work}
\label{sec:conclusion}
We introduced Latent Recurrent Transformer (LRT), which reuses a high-level source-layer state $\mathbf{m}_{t-1}=\mathbf{h}^{\ell_{\mathrm{src}}}_{t-1}$ as recurrent memory for the next token while preserving one normal forward pass per generated token.
LRT shifts recurrent refinement from inference-time extra computation to more parallelizable pretraining. Future work could improve the training efficiency of interleaved parallel training, and broader downstream evaluation, especially on tasks that may benefit from cross-token recurrent computation, such as mathematical reasoning, code generation, and long-context question answering. Finally, LRT could be combined with complementary forms of additional computation, such as latent thought tokens.

\section*{Limitations}
This work is an architectural and training-strategy exploration evaluated primarily on language-modeling and in-context learning metrics under nanochat-style pretraining settings. The experiments do not yet cover larger production-scale models, broader downstream tasks such as code generation or mathematical reasoning. Interleaved parallel training also introduces additional training compute relative to a standard transformer update, and its hardware efficiency may depend on implementation details.

\section*{Acknowledgments}
This work was supported in part by NSF IIS2404180, Institute of Information \& communications Technology Planning \& Evaluation (IITP) grant funded by the Korea government (MSIT) (No. 2022-0-00871, Development of AI Autonomy and Knowledge Enhancement for AI Agent Collaboration), and Electronics and Telecommunications Research Institute (ETRI) grant (26CB1200, Development and Application of Science-Specialized Multimodal Foundation Models). We used ChatGPT to assist with language editing and LaTeX/camera-ready preparation. All technical content, experimental results, claims, and references were reviewed and verified by the authors.
\bibliography{main}

\clearpage
\appendix
\setcounter{table}{0}
\setcounter{figure}{0}
\renewcommand{\thetable}{\Alph{section}.\arabic{table}}
\renewcommand{\thefigure}{\Alph{section}.\arabic{figure}}


\section{Implementation Details}
\label{app:implementation}

\paragraph{Backbone.}
We follow the nanochat training stack~\citep{karpathy2025nanochat}. The baseline is a pre-norm decoder-only Transformer~\citep{vaswani2017attention,radford2019language,brown2020language} with untied input embeddings and LM head, parameter-free RMSNorm~\citep{zhang2019root}, RoPE positional encodings with base $10{,}000$~\citep{su2024roformer}, QK normalization, and standard multi-head self-attention. Attention uses a tiled sliding-window pattern with an ``SSSL'' schedule, where the final layer always uses full context. MLPs use two bias-free linear layers with hidden width $4d$ and a squared-ReLU activation, and the model uses no biases or dropout. Logits are soft-capped as $\mathbf{z} \leftarrow c \tanh(\mathbf{z}/c)$ with $c=15$, following recent large-model practice~\citep{team2024gemma}. The vocabulary is padded to $32{,}768$ tokens for efficient matrix multiplication.

\paragraph{Residual stream and value embeddings.}
Each block applies learnable residual scaling and a skip connection to the initial embedding. Nanochat value embeddings are mixed into attention values on all layers, through an input-dependent gate. These baseline components are kept fixed across the baseline and LRT variants, so the comparison isolates the effect of the LRT recurrent memory pathway.

\paragraph{Model scaling.}
We follow nanochat's constant-aspect-ratio scaling rule, setting $d=64L$ with head dimension $128$. We report results at 20L ($L=20$, $d=1280$, 10 heads) and 24L ($L=24$, $d=1536$, 12 heads). 

\paragraph{LRT variants.}
The default LRT variant, LRT-shared, shares recurrent KV projection matrices across layers and adds $0.3\%$ parameters. LRT-layerwise uses separate recurrent projections per layer and adds $4.8\%$ parameters for 20L and $5.4\%$ for 24L. Unless otherwise specified, LRT uses combined KV Projection and Residual Injection, dual gating, zero memory initialization, source-layer recurrent memory selected by validation ablation, and interleaved parallel training with $S=2$.

\paragraph{Data.}
We pretrain on FineWeb-Edu 100BT~\citep{lozhkovfineweb} using the pre-shuffled nanochat release~\citep{karpathy2025nanochat}. The final shard is held out as validation and the remaining shards are used for training. Documents are tokenized on the fly with the nanochat BPE tokenizer with vocabulary size $32{,}768$. We train with MuonAdamW following nanochat defaults~\citep{jordan2024muon}, using a global batch of $2^{19}$ tokens, sequence length $2048$, bfloat16 mixed precision, and 8 H100 GPUs. We use no warmup and linearly warm down over the final half of training.

\paragraph{Optimization.}

We optimize all models with MuonAdamW, following the nanochat training recipe~\citep{karpathy2025nanochat} and Muon optimizer setup~\citep{jordan2024muon}. Muon is applied to all two-dimensional weight matrices, including attention projections, MLP matrices, and LRT projection matrices when present. AdamW is used for token embeddings, the LM head, value embeddings, and scalar parameters such as residual and input-skip coefficients, layer scales, and LRT gating or scaling parameters.

At the reference width $d=768$ and reference batch size $2^{19}$ tokens, we use matrix learning rate $0.02$ for Muon, embedding and value-embedding learning rate $0.3$, LM-head learning rate $0.004$, and scalar learning rate $0.5$ (with the residual coefficients further scaled down by $0.01$). Adam uses $\beta=(0.8,0.95)$, except the input-skip coefficients, which use $\beta=(0.96,0.95)$. The embedding, value-embedding, and LM-head learning rates are scaled by $(d/768)^{-1/2}$, and all learning rates are scaled by $\sqrt{\mathrm{batch}/2^{19}}$ when the global batch size differs from the reference batch. We use no warmup and linearly warm down the learning rate to zero over the final half of training. Muon momentum is linearly increased from $0.85$ to $0.95$ over the first $300$ steps, and weight decay is scaled by $(12/L)^2$ and linearly annealed to zero over training.

\paragraph{Evaluation.}
Validation BPB is computed on the held-out FineWeb-Edu shard~\citep{lozhkovfineweb} and normalized by UTF-8 byte count. CORE evaluation follows the DCLM evaluation suite~\citep{li2024datacomp} as implemented in nanochat. Full CORE task details are provided in Appendix~\ref{app:core_details}.


\section{Compute Budget and Full Scaling Results}
\label{app:scaling_tables}

\begin{table*}[t]
\centering
\scriptsize
\setlength{\tabcolsep}{3.2pt}
\begin{tabular}{@{}llccrrrrrrr@{}}
\toprule
\textbf{Model}
& \textbf{Schedule}
& \textbf{Training Cost}
& \textbf{$\Delta$ Param.}
& \multicolumn{7}{c}{\textbf{Raw tokens per parameter, $R$}} \\
\cmidrule(lr){5-11}
& & & & \textbf{5} & \textbf{10} & \textbf{20}
& \textbf{40} & \textbf{60} & \textbf{80} & \textbf{120} \\
\midrule
20L Base
& -- & $1\times$ & --
& 0.789 & 0.767 & 0.751 & 0.739 & 0.734 & 0.731 & 0.726 \\

20L LRT-shared
& Interleaved ($S=2$) & $\sim2\times$ & $0.3\%$
& 0.777 & 0.754 & 0.739 & 0.727 & 0.721 & 0.717 & -- \\

20L LRT-layerwise
& Interleaved ($S=2$) & $\sim2\times$ & $4.8\%$
& 0.775 & 0.752 & 0.735 & 0.725 & 0.719 & 0.715 & -- \\
\midrule
24L Base
& -- & $1\times$ & --
& 0.753 & 0.733 & 0.719 & 0.709 & 0.703 & 0.699 & 0.694 \\

24L LRT-shared
& Interleaved ($S=2$) & $\sim2\times$ & $0.3\%$
& 0.742 & 0.722 & 0.706 & 0.695 & 0.689 & 0.685 & -- \\

24L LRT-layerwise
& Interleaved ($S=2$) & $\sim2\times$ & $5.4\%$
& 0.739 & 0.720 & 0.704 & 0.693 & 0.687 & 0.683 & -- \\
\bottomrule
\end{tabular}
\caption{
Full BPB scaling results. All LRT results in this table use interleaved
parallel training with $S=2$; full-sequence refinement configurations are
reported separately. Cost denotes approximate training compute relative to a
standard Transformer update, and $\Delta$ Param.\ denotes additional
trainable parameters relative to the corresponding baseline. LRT-shared uses
recurrent projections shared across layers, whereas LRT-layerwise uses
separate projections for each layer. Because interleaved training costs
approximately $2\times$, an LRT model trained at raw ratio $R$ is plotted at
$2R$ in baseline-equivalent compute comparisons. Lower BPB is better.
}
\label{tab:bpb_scaling_full}
\end{table*}

\begin{table*}[t]
\centering
\scriptsize
\setlength{\tabcolsep}{3.2pt}
\begin{tabular}{@{}llccrrrrrrr@{}}
\toprule
\textbf{Model}
& \textbf{Schedule}
& \textbf{Training Cost}
& \textbf{$\Delta$ Param.}
& \multicolumn{7}{c}{\textbf{Raw tokens per parameter, $R$}} \\
\cmidrule(lr){5-11}
& & & & \textbf{5} & \textbf{10} & \textbf{20}
& \textbf{40} & \textbf{60} & \textbf{80} & \textbf{120} \\
\midrule
20L Base
& -- & $1\times$ & --
& 0.217 & 0.242 & 0.252 & 0.261 & 0.267 & 0.271 & 0.276 \\

20L LRT-shared
& Interleaved ($S=2$) & $\sim2\times$ & $0.3\%$
& 0.226 & 0.251 & 0.263 & 0.274 & 0.280 & 0.285 & -- \\

20L LRT-layerwise
& Interleaved ($S=2$) & $\sim2\times$ & $4.8\%$
& 0.228 & 0.254 & 0.266 & 0.277 & 0.284 & 0.289 & -- \\
\midrule
24L Base
& -- & $1\times$ & --
& 0.233 & 0.276 & 0.291 & 0.304 & 0.309 & 0.312 & 0.316 \\

24L LRT-shared
& Interleaved ($S=2$) & $\sim2\times$ & $0.3\%$
& 0.251 & 0.283 & 0.301 & 0.313 & 0.317 & 0.320 & -- \\

24L LRT-layerwise
& Interleaved ($S=2$) & $\sim2\times$ & $5.4\%$
& 0.254 & 0.285 & 0.303 & 0.315 & 0.319 & 0.322 & -- \\
\bottomrule
\end{tabular}
\caption{
Full CORE scaling results. All LRT results in this table use interleaved
parallel training with $S=2$; full-sequence refinement configurations are
reported separately. Cost denotes approximate training compute relative to a
standard Transformer update, and $\Delta$ Param.\ denotes additional
trainable parameters relative to the corresponding baseline. LRT-shared uses
recurrent projections shared across layers, whereas LRT-layerwise uses
separate projections for each layer. Because interleaved training costs
approximately $2\times$, an LRT model trained at raw ratio $R$ is plotted at
$2R$ in baseline-equivalent compute comparisons. Higher CORE is better.
}
\label{tab:core_scaling_full}
\end{table*}

We report full BPB and CORE scaling results in Tables~\ref{tab:bpb_scaling_full} and~\ref{tab:core_scaling_full}. 
The training budget is set by the tokens-per-parameter ratio $R$: a model with $N$ trainable parameters is trained on approximately $R \times N$ tokens. 
In the main figures, baseline points are plotted at baseline-equivalent compute $R$, while LRT points are plotted at $2R$ because interleaved parallel training costs approximately two standard transformer updates.

\section{Additional Architecture Ablations}
\label{app:architecture_ablation}

We provide additional ablations of the recurrent memory-injection mechanism on the 20L model at ratio $R=10$. These experiments compare three design choices: whether memory is injected through keys, values, or both; whether recurrent KV features are added to or replace local KV features; and whether KV Projection is combined with Residual Injection.

\begin{table}[t]
\centering
\small
\setlength{\tabcolsep}{4pt}
\begin{tabularx}{\columnwidth}{Xc}
\toprule
\textbf{Architecture} & \textbf{BPB} $\downarrow$ \\
\midrule
Baseline / no memory & 0.767 \\
\midrule
Value-only projection & 0.756 \\
Key-only projection & 0.762 \\
KV Projection & 0.755 \\
KV replace, early 1/3 layers & 0.760 \\
\midrule
Residual Injection & 0.756 \\
KV Projection + Residual Injection & \textbf{0.754} \\
\bottomrule
\end{tabularx}
\caption{
Detailed architecture ablations on the 20L model at ratio $R=10$. BPB is lower better.
``KV replace'' denotes replacing local KV features with recurrent projections in the early one-third of layers.
Injecting recurrent memory into values is stronger than injecting it into keys alone, suggesting that recurrent memory is especially useful as attention content. Adding recurrent key-value features to local features outperforms replacing local KV features, indicating that recurrent memory is more useful as an auxiliary pathway than as a substitute for local token features. Combining KV Projection with Residual Injection performs best, suggesting that attention-level and residual-stream memory pathways are complementary.
}
\label{tab:memory_injection_ablation_full}
\end{table}

The results support three design choices. First, value-only projection is stronger than key-only projection, while full KV Projection performs best among KV-only variants; we therefore project recurrent memory into both keys and values. Second, additive KV Projection outperforms replacing local KV features, suggesting that recurrent memory should augment rather than substitute local token features. Third, combining KV Projection with Residual Injection gives the best BPB, indicating that attention-level and residual-stream pathways expose recurrent memory to complementary parts of the transformer computation.

\section{Training Strategy Ablations}
\label{app:training_strategy_ablations}

\subsection{Full-Sequence Refinement}
\label{app:parallel_training_analysis}

\paragraph{Training details.}
When $S=1$, each refinement forward recomputes the entire sequence in
parallel. We use $K$ to denote the number of refinement forwards following
initialization. Starting from the initialization buffer
$\mathcal{B}^{(0)}$, refinement forward $k$ reads
$\mathcal{B}^{(k-1)}$ and writes the updated states to
$\mathcal{B}^{(k)}$. All positions are updated synchronously, so states
produced in one refinement forward become available only in the next forward.

By default, we apply language-modeling supervision to the initialization and
every refinement forward:
\begin{equation}
    \mathcal{L}
    =
    \frac{1}{K+1}
    \left(
        \mathcal{L}_{\mathrm{init}}
        +
        \sum_{k=1}^{K}\mathcal{L}_{\mathrm{full}}^{(k)}
    \right).
\end{equation}
We backpropagate jointly through all $K+1$ forwards without detaching the
intermediate buffers. We also evaluate a weighted variant,
$\mathcal{L}=\sum_{k=0}^{K}w_k\mathcal{L}^{(k)}/\sum_{k=0}^{K}w_k$,
which places greater weight on later refinement forwards.
For the weighted-loss variant with $K=2$, we assign weights $0.1$, $0.3$, and $0.6$ to the initialization, first-refinement, and second-refinement losses, respectively.

\paragraph{Number of refinement forwards.}
Table~\ref{tab:full_sequence_refinement} studies the effect of increasing
$K$. A single refinement forward reaches 0.753 BPB. Increasing $K$ further
improves BPB, but the gains diminish and saturate around three refinement
forwards. The weighted-loss variant reaches the best BPB of 0.748 with
$K=2$, suggesting that emphasizing the more refined representations can be
beneficial.

\begin{table*}[t]
\centering
\small
\setlength{\tabcolsep}{5pt}
\begin{tabular}{lccccccc}
\toprule
\textbf{Training setting}
& \textbf{$K$}
& \textbf{Init}
& \textbf{Refine 1}
& \textbf{Refine 2}
& \textbf{Refine 3}
& \textbf{Refine 4}
& \textbf{Decoding} \\
\midrule
Transformer baseline
& 0 & 0.767 & -- & -- & -- & -- & 0.767 \\
Full-sequence refinement
& 1 & 0.762 & 0.753 & -- & -- & -- & 0.753 \\
Full-sequence refinement
& 2 & 0.763 & 0.751 & 0.751 & -- & -- & 0.751 \\
Full-sequence refinement, weighted loss
& 2 & 0.773 & 0.750 & \textbf{0.748} & -- & -- & \textbf{0.748} \\
Full-sequence refinement
& 3 & 0.766 & 0.751 & 0.749 & 0.749 & -- & 0.749 \\
Full-sequence refinement
& 4 & 0.769 & 0.753 & 0.751 & 0.749 & 0.749 & 0.749 \\
\bottomrule
\end{tabular}
\caption{
Ablation on the number of full-sequence refinement forwards using the 20-layer
model. Each intermediate column reports BPB after the corresponding forward,
and ``Decoding'' reports BPB under recurrent autoregressive inference. Lower
BPB is better. Additional refinement initially improves performance, but the
gain saturates as $K$ increases. The weighted-loss configuration uses weights $(0.1,0.3,0.6)$ for the
initialization and two refinement forwards.
}
\label{tab:full_sequence_refinement}
\end{table*}

\subsection{Interleaved versus Full-Sequence Refinement}
\label{app:interleaved_full_comparison}

Table~\ref{tab:interleaved_full_scaling} compares interleaved and
full-sequence refinement across pretraining data budgets. At matched
approximately $2\times$ training compute, interleaved refinement with $S=2$
slightly outperforms full-sequence refinement with $K=1$ at $R=5$ and
$R=10$, while the two schedules perform identically from $R=20$ onward.
This suggests that the immediate write-back between interleaved subsets may
be most useful in the low-data regime, whereas the difference between the
two schedules disappears with more training data.

Additional full-sequence refinement forwards provide consistent but
incremental improvements. Increasing $K$ from 1 to 3 reduces BPB across all
data budgets, with training cost increasing linearly from approximately
$2\times$ to $4\times$. We therefore use interleaved refinement with $S=2$
as our default configuration, while full-sequence refinement provides a
simple way to trade additional training compute for improved performance.

\begin{table*}[t]
\centering
\scriptsize
\setlength{\tabcolsep}{3.5pt}
\begin{tabular}{@{}llccrrrrrrr@{}}
\toprule
\textbf{Model}
& \textbf{Schedule}
& \textbf{Cost}
& \textbf{$\Delta$ Param.}
& \multicolumn{7}{c}{\textbf{Raw tokens per parameter, $R$}} \\
\cmidrule(lr){5-11}
& & & & \textbf{5} & \textbf{10} & \textbf{20}
& \textbf{40} & \textbf{60} & \textbf{80} & \textbf{120} \\
\midrule
Transformer
& -- & $1\times$ & -- 
& 0.789 & 0.767 & 0.751 & 0.739 & 0.734 & 0.731 & 0.726 \\

LRT-layerwise
& Interleaved ($S=2$) & $\sim2\times$ & $4.8\%$
& 0.775 & 0.752 & 0.735 & 0.725 & 0.719 & 0.715 & -- \\

LRT-layerwise
& Full-seq.\ ($K=1$) & $\sim2\times$ & $4.8\%$
& 0.777 & 0.753 & 0.735 & 0.725 & 0.719 & 0.715 & -- \\

LRT-layerwise
& Full-seq.\ ($K=2$) & $\sim3\times$ & $4.8\%$
& 0.774 & 0.751 & 0.733 & 0.723 & 0.717 & 0.713 & -- \\

LRT-layerwise
& Full-seq.\ ($K=3$) & $\sim4\times$ & $4.8\%$
& 0.772 & 0.749 & 0.731 & 0.721 & 0.716 & 0.711 & -- \\
\bottomrule
\end{tabular}
\caption{
Comparison of interleaved and full-sequence refinement across pretraining
data budgets using the same 20-layer backbone. At matched approximately
$2\times$ training compute, interleaved refinement with $S=2$ slightly
outperforms full-sequence refinement with $K=1$ in the low-data regime and
matches it as $R$ increases. Additional full-sequence refinement forwards
provide incremental gains at linearly increasing cost. Here, $S$ denotes the
number of interleaved subsets, while $K$ denotes the number of full-sequence
refinement forwards following initialization. Cost is approximate training
compute relative to the Transformer baseline, and $\Delta$ Param.\ denotes
additional trainable parameters. Lower BPB is better.
}
\label{tab:interleaved_full_scaling}
\end{table*}

\subsection{Comparison with Exact Sequential Training}
\label{app:exact_recurrent_training}

Exact recurrent training follows the inference-time dependency exactly,
processing each token only after obtaining the recurrent state of its
predecessor. This removes sequence-level parallelism and requires $T$
dependent forward stages for a length-$T$ sequence, making it prohibitively
slow at our main scale. We therefore conduct a controlled comparison on
smaller 8- and 12-layer models trained for 4,000 steps with sequence length
256.

As shown in Table~\ref{tab:exact_recurrent_training}, exact sequential
training performs best, but interleaved training remains within 0.004 BPB at
both depths. Relative to the Transformer baseline, interleaved training
recovers 75.0\% and 76.5\% of the improvements achieved by exact recurrence
for the 8- and 12-layer models, respectively, while reducing the number of
dependent forward stages from 256 to 3. Full-sequence refinement with $K=3$
further reduces the gap to 0.002 BPB and recovers approximately 88\% of the
exact-recurrence improvement, although it uses more training compute than the
default interleaved schedule. These results suggest that parallel refinement
captures most of the benefit of exact token-level recurrence while remaining
substantially more parallel.

\begin{table}[t]
\centering
\scriptsize
\setlength{\tabcolsep}{4pt}
\begin{tabular}{@{}lccc@{}}
\toprule
\textbf{Training strategy}
& \textbf{Stages}
& \textbf{8L BPB} $\downarrow$
& \textbf{12L BPB} $\downarrow$ \\
\midrule
Transformer
& 1 & 1.082 & 1.040 \\
LRT, interleaved ($S=2$)
& 3 & 1.070 & 1.027 \\
LRT, full-seq.\ ($K=3$)
& 4 & 1.068 & 1.025 \\
LRT, exact sequential
& 256 & \textbf{1.066} & \textbf{1.023} \\
\bottomrule
\end{tabular}
\caption{
Comparison with exact token-by-token recurrent training on smaller models
trained for 4,000 steps with sequence length 256. ``Stages'' denotes the
number of dependency-ordered forward stages per sequence. Exact recurrence
achieves the lowest BPB but requires one stage per token, whereas the parallel
refinement schedules recover most of its improvement using only three or four
stages. Full-sequence refinement with $K=3$ uses more training compute than
interleaved refinement with $S=2$ and is included as an accuracy--compute
reference rather than a matched-compute comparison. Lower BPB is better.
}
\label{tab:exact_recurrent_training}
\end{table}

\subsection{Chunked Training}
\label{app:chunked_training_ablation}

Chunked training is a low-token-compute approximation to recurrent training. It splits a sequence into contiguous chunks and processes chunks sequentially, carrying recurrent memory across chunk boundaries. Within each chunk, computation remains parallel. Thus, in ideal token-compute terms, each token is processed once and the cost is close to a standard transformer update.

However, chunked training gives a coarse approximation to token-level recurrence. Recurrent memory is updated only across chunk boundaries rather than after every token, so tokens inside the same chunk do not receive recurrent memory from their immediate predecessors. Smaller chunks create more frequent recurrent updates, but require more sequential chunk forwards; larger chunks improve parallelism, but make the recurrent signal sparser.

This trade-off is visible in Table~\ref{tab:chunked_training_ablation}. With chunk size $C=64$, chunked training slightly improves over the baseline, but remains far behind interleaved parallel training. With $C=256$, the recurrent signal becomes too sparse and the result is nearly identical to the baseline. Moreover, small chunks are inefficient in wall-clock training: with sequence length $2048$, $C=64$ requires 32 sequential chunk forwards per sequence, while $C=256$ still requires 8. This makes chunked training difficult to scale efficiently without specialized systems support.

\begin{table}[t]
\centering
\small
\setlength{\tabcolsep}{4pt}
\begin{tabularx}{\columnwidth}{Xccc}
\toprule
\textbf{Training strategy} & \textbf{Ideal cost} & \textbf{R10} & \textbf{R20} \\
\midrule
Baseline & $1\times$ & 0.767 & 0.751 \\
Interleaved training & $\sim 2\times$ & \textbf{0.754} & \textbf{0.739} \\
Chunked ($C=64$) & $\sim 1\times$ & 0.765 & 0.750 \\
Chunked ($C=256$) & $\sim 1\times$ & 0.767 & 0.751 \\
\bottomrule
\end{tabularx}
\caption{
Chunked training ablation on the 20L model. BPB is lower better. Chunked training has low ideal token compute, but its recurrent signal is coarse because memory is propagated only across chunk boundaries. Smaller chunks provide more frequent recurrent updates but require many sequential chunk forwards, while larger chunks improve parallelism at the cost of sparser recurrence. Interleaved training gives stronger gains by providing recurrent-memory-aware refinement for all positions.
}
\label{tab:chunked_training_ablation}
\end{table}

\subsection{Measured Efficiency and Parallel Prefill}
\label{app:efficiency}

We measure the training and serving efficiency of LRT against a Loop
Transformer with a matched backbone configuration. All methods use
PyTorch~2.9, FlashAttention-3, BF16 precision, and H100-80GB GPUs. Training is
measured on eight H100 GPUs with 524,288 tokens per optimization step. Serving
is measured on a single H100 with batch size 128, a prompt length of 1,024
tokens, and 1,024 generated tokens per sequence. Each entry in
Table~\ref{tab:measured_efficiency} reports throughput, latency, and peak
memory relative to the Transformer baseline. Higher throughput and lower
latency and memory are better.

\begin{table}[t]
\centering
\small
\setlength{\tabcolsep}{8pt}
\renewcommand{\arraystretch}{1.15}
\begin{tabular}{@{}lcc@{}}
\toprule
\textbf{Stage}
& \shortstack{\textbf{Loop Transformer}\\\textbf{(2 loops)}}
& \textbf{LRT-shared} \\
\midrule
Training
& $0.45 / 2.24 / 1.09$
& $0.36 / 2.96 / 1.19$ \\
Prefill
& $0.44 / 2.26 / 1.70$
& $1.00 / 1.00 / 1.00$ \\
Decoding
& $0.44 / 2.27 / 1.81$
& $0.92 / 1.09 / 1.00$ \\
\bottomrule
\end{tabular}
\caption{
Measured efficiency relative to the Transformer baseline. Each entry reports
\emph{throughput / latency / peak memory}. The Loop Transformer uses two
loops. LRT training uses interleaved parallel training with $S=2$, while the
reported LRT prefill entry uses one-pass prefill. Measurements for interleaved
prefill are reported below.
}
\label{tab:measured_efficiency}
\end{table}

For reference, the absolute Transformer measurements are 1,183,622 tokens/s
and 442.9\,ms per training step, 198,389 tokens/s and 0.66\,s per prefill
batch, and 11,934 tokens/s and 10.73\,ms per decoding step. For LRT, the
measured training overhead is higher than the ideal $2\times$ token-compute
estimate because the initialization and subset-refinement forwards are
executed sequentially and the smaller subset forwards have lower hardware
utilization.

\paragraph{Parallel prefill.}
LRT supports both training-matched interleaved prefill and one-pass prefill.
Interleaved prefill applies the same refinement schedule used during training
and obtains $0.35\times$ throughput, $2.98\times$ latency, and $1.11\times$
peak memory relative to the Transformer. One-pass prefill instead has
baseline-equivalent throughput, latency, and peak memory.

In our evaluation, the choice between these strategies has little effect on
continuation quality. When BPB is evaluated over the 1,024 tokens following a
1,024-token prompt, both remain within $0.001$ BPB of exact sequential
prefill: interleaved prefill differs by $+0.00002$, while one-pass prefill
differs by $+0.00082$. Thus, interleaved prefill more closely reproduces exact
recurrence, but both strategies provide nearly identical continuation quality
in this setting. The prefill strategy can therefore be selected according to
the desired latency quality trade-off. In either case, the source-layer state
at the final prompt position initializes recurrent decoding.

During decoding, LRT performs one forward per generated token and stores only
one additional $d$-dimensional recurrent state per sequence. Under our
measurement setting, this results in $0.92\times$ throughput,
$1.09\times$ latency, and no measurable increase in peak memory relative to
the Transformer.

\subsection{Number of Interleaved Subsets}
\label{app:interleaved_subset_ablation}

Interleaved Training refines $S$ disjoint subsets of positions after one full-sequence initialization pass. Increasing $S$ creates a longer sparse recurrent refinement chain: later subsets can consume recurrent states updated by more earlier subsets. However, because only one subset is refreshed at each refinement step, larger $S$ also means later subsets rely on a buffer that is only partially updated from the initialization pass. Thus, increasing $S$ may not necessarily improve the approximation to the full token-level recurrent chain.

We evaluate $S \in \{2,4,8\}$ on the 20L model at ratio $R=10$. Since the subsets together cover the sequence exactly once, the ideal token compute is approximately unchanged across $S$: one full initialization pass plus one effective refinement pass, or about $2\times$ a standard transformer update. In practice, larger $S$ can introduce more sequential refinement steps and may reduce hardware efficiency, even if the ideal token count is unchanged.

\begin{table}[t]
\centering
\small
\begin{tabular}{lcc}
\toprule
\textbf{Setting} & \textbf{Training cost} & \textbf{BPB} $\downarrow$ \\
\midrule
Baseline / no memory & $1\times$ & 0.767 \\
LRT, $S=2$ & $\sim 2\times$ & 0.754 \\
LRT, $S=4$ & $\sim 2\times$ & 0.754 \\
LRT, $S=8$ & $\sim 2\times$ & 0.754 \\
\bottomrule
\end{tabular}
\caption{
Ablation on the number of interleaved subsets $S$ on the 20L model at ratio $R=10$. Lower BPB is better. Increasing $S$ creates a longer sparse refinement chain, but does not improve BPB in this setting. We therefore use $S=2$ by default, which is the simplest and most parallel option that matches the best observed BPB.
}
\label{tab:interleaved_subset_ablation}
\end{table}

The results show that increasing the number of interleaved subsets does not improve BPB: all LRT variants obtain 0.754. This suggests that, under our current approximation, the benefit mainly comes from giving each token one recurrent-memory-aware refinement step rather than from increasing the number of sparse write-back stages. We therefore use $S=2$ as the default setting because it provides the same BPB as larger $S$ while requiring fewer sequential refinement stages and preserving more parallelism.

\section{Non-Commutative State Tracking}
\label{app:state_tracking}

We evaluate LRT on $\mathrm{A}_5$ prefix group multiplication, a
non-commutative state-tracking task over a group of order 60. At each
position, the model must predict the cumulative group product of the observed
prefix. We train depth-4, $d=256$ Transformer and LRT models under identical
settings using length-64 training sequences. LRT uses interleaved parallel
training with $S=2$.

We report \emph{pos@90}, defined as the largest position $p$ such that
accuracy remains at least $90\%$ at every position from 1 through $p$.
As shown in Table~\ref{tab:state_tracking}, the Transformer reaches
$\mathrm{pos@90}=5$, whereas LRT reaches $\mathrm{pos@90}=24$, corresponding
to a $4.8\times$ longer reliable state-tracking horizon. LRT additionally
maintains $100\%$ accuracy through position 16.

\begin{table}[t]
\centering
\small
\setlength{\tabcolsep}{5pt}
\begin{tabular}{@{}lccc@{}}
\toprule
& \multicolumn{3}{c}{\textbf{Evaluation length}} \\
\cmidrule(lr){2-4}
\textbf{Model}
& \textbf{64}
& \textbf{128}
& \textbf{256} \\
\midrule
Transformer
& 5 & 5 & 5 \\
LRT
& \textbf{24} & \textbf{24} & \textbf{24} \\
\bottomrule
\end{tabular}
\caption{
$\mathrm{pos@90}$ on $\mathrm{A}_5$ prefix group multiplication. Both models
are trained on length-64 sequences; LRT uses interleaved training with
$S=2$. LRT extends the reliable state-tracking horizon from 5 to 24 positions,
and the result is unchanged across evaluation lengths. Higher is better.
}
\label{tab:state_tracking}
\end{table}

The position-wise accuracy curves are identical over their shared prefixes
when evaluated at lengths 64, 128, and 256. Thus, the measured reliable
horizon is not an artifact of the selected evaluation length. As a control,
the same Transformer configuration reaches $99.6\%$ accuracy on prefix
multiplication over the cyclic group $\mathbb{Z}_{60}$, which has the same
number of group elements as $\mathrm{A}_5$ but is commutative. This result
indicates that the difficulty of $\mathrm{A}_5$ is not simply caused by the
60-way output space or the evaluation pipeline, and that LRT provides a
substantial advantage on non-commutative state tracking.

\section{Statistical Reliability}
\label{app:statistical_reliability}

All BPB and CORE results reported in this paper are means over three
independent runs. The observed improvements are consistently larger than the
corresponding run-to-run variation. These gains are also obtained over a
strong nanochat baseline, for which common architectural additions such as
gated attention, layer scaling, and residual mixing provide little or no
improvement (Table~\ref{tab:baseline_negative_ablations}).

Because interleaved LRT training uses approximately $2\times$ the compute of
a standard Transformer update, a matched-compute comparison pairs LRT trained
at raw tokens-per-parameter ratio $R$ with the baseline trained at ratio
$2R$. Table~\ref{tab:statistical_reliability} reports results at three such
baseline-equivalent compute budgets.

\begin{table*}[t]
\centering
\small
\setlength{\tabcolsep}{7pt}
\begin{tabular}{@{}ccccc@{}}
\toprule
\textbf{Equivalent compute}
& \textbf{Base BPB} $\downarrow$
& \textbf{LRT BPB} $\downarrow$
& \textbf{Base CORE} $\uparrow$
& \textbf{LRT CORE} $\uparrow$ \\
\midrule
40
& $0.7394 \pm 0.0002$
& $\mathbf{0.7387 \pm 0.0002}$
& $0.2608 \pm 0.0007$
& $\mathbf{0.2627 \pm 0.0008}$ \\
80
& $0.7308 \pm 0.0002$
& $\mathbf{0.7272 \pm 0.0002}$
& $0.2707 \pm 0.0007$
& $\mathbf{0.2742 \pm 0.0007}$ \\
120
& $0.7261 \pm 0.0001$
& $\mathbf{0.7213 \pm 0.0001}$
& $0.2760 \pm 0.0006$
& $\mathbf{0.2803 \pm 0.0006}$ \\
\bottomrule
\end{tabular}
\caption{
Matched-compute BPB and CORE results, reported as mean $\pm$ standard
deviation over three independent runs. At equivalent compute $C$, the
Transformer baseline is trained at raw ratio $C$, while LRT is trained at raw
ratio $C/2$ because its interleaved training schedule costs approximately
$2\times$. LRT consistently improves both metrics by margins larger than the
observed run-to-run variation.
}
\label{tab:statistical_reliability}
\end{table*}

Across all three matched-compute budgets, LRT consistently improves both BPB
and CORE. The improvement grows with compute and remains larger than the
reported standard deviations, supporting that the observed gains are
reproducible rather than explained by run-to-run variation.

\section{Baseline Strength and Negative Ablations}
\label{app:baseline_ablations}

To contextualize the BPB improvements reported in the main paper, we evaluate several generic architectural additions on the same nanochat-style transformer backbone. These variants add extra flexibility, such as residual scaling, gated attention, input residual mixing, or value-side residual paths, but do not introduce the LRT recurrent memory pathway. The goal is to verify that the gains from LRT are not simply explained by adding small modules or gates to the baseline.

\begin{table}[ht]
\centering
\small
\setlength{\tabcolsep}{4pt}
\begin{tabularx}{\columnwidth}{Xc}
\toprule
\textbf{Variant} & \textbf{BPB} $\downarrow$ \\
\midrule
Vanilla baseline & 0.784 \\
Layer-scaled attn./MLP & 0.782 \\
Gated attention (GA) & 0.782 \\
Input residual mixing (IRM) & 0.779 \\
Dense residual mixing & 0.783 \\
Value residual & 0.779 \\
Projected value residual & 0.780 \\
Value embedding (VE) & 0.770 \\
VE + IRM (final baseline) & 0.767 \\
VE + IRM + GA & 0.767 \\
\bottomrule
\end{tabularx}
\caption{
Baseline-strength and negative ablations on the 20L model. BPB is lower better.
VE denotes value embedding, IRM denotes input residual mixing, and GA denotes gated attention.
Generic additions such as gated attention, layer scaling, and residual mixing yield only marginal improvements, and adding gated attention on top of the final baseline does not further improve BPB. This suggests that the gains reported for LRT are not simply due to adding extra parameters, gates, or residual pathways, but instead come from the recurrent memory mechanism.
}
\label{tab:baseline_negative_ablations}
\end{table}

The strongest non-recurrent baseline combines value embeddings with input residual mixing, reaching 0.767 BPB. Adding another generic gate on top of this baseline does not improve performance. We therefore use this strong nanochat-style configuration as the baseline in the main experiments, and compare LRT against it rather than against the weaker vanilla transformer.

For reference, layer-scaled attention/MLP uses 
$x \leftarrow x + \alpha \mathrm{Attn}(x) + \beta \mathrm{MLP}(x)$; 
gated attention uses $x \leftarrow x + \alpha \mathrm{Attn}(x)$; 
input residual mixing uses $x \leftarrow \alpha x + \beta x_0$; 
and dense residual mixing additionally incorporates earlier residual states.

\section{Training Algorithm}
We show the full interleaved parallel training algorithm in Alg.~\ref{app:training_algorithm}.

\label{app:training_algorithm}

  \begin{algorithm}[t]
  \caption{Interleaved Parallel Training Step}
  \label{alg:interleaved}
  \begin{algorithmic}[1]
  \STATE Inputs: token sequence $x_{1:T}$, subset count $S$, partition
         $\{\mathcal{I}_s\}_{s=1}^{S}$
  \STATE $\mathcal{B}^{(0)} \leftarrow \mathrm{Forward}_{\text{full}}(x_{1:T})$
         \hfill // init forward: parallel over $T$
  \STATE compute $\mathcal{L}_{\mathrm{init}}$ on $\mathcal{B}^{(0)}$
  \FOR{$s = 1$ to $S$}
     \STATE $\mathcal{B}^{(s)} \leftarrow \mathrm{Refine}(\mathcal{I}_s,
            \mathcal{B}^{(s-1)})$
            \hfill // compute $|\mathcal{I}_s|$ positions; write back
     \STATE compute $\mathcal{L}_{\mathcal{I}_s}$ on refined positions
  \ENDFOR
  \STATE $\mathcal{L} \leftarrow \tfrac{1}{S+1}\!\big(
         \mathcal{L}_{\mathrm{init}} + \sum_s \mathcal{L}_{\mathcal{I}_s}\big)$
  \STATE backpropagate $\mathcal{L}$ through all $S{+}1$ forwards
  \end{algorithmic}
  \end{algorithm}

\section{CORE Evaluation Details}
\label{app:core_details}

CORE is computed over 22 in-context learning tasks from the DCLM evaluation suite~\citep{li2024datacomp}. The tasks are grouped into five categories: language understanding, world knowledge, commonsense reasoning, symbolic problem solving, and reading comprehension. Each task contributes a centered score, $\mathrm{centered}=(a-r)/(1-r)$, where $a$ is the task accuracy and $r$ is the random baseline. The final CORE metric is the mean centered score across all tasks. In the implementation, random baselines stored as percentages are converted to probabilities before centering.

\begin{table*}[t]
\centering
\small
\begin{tabular}{llcl}
\toprule
\textbf{Category} & \textbf{Task} & \textbf{Shots} & \textbf{Type} \\
\midrule
Language understanding & hellaswag\_zeroshot & 0 & multiple choice \\
Language understanding & hellaswag & 10 & multiple choice \\
Language understanding & lambada\_openai & 0 & language modeling \\
Language understanding & winograd & 0 & schema \\
Language understanding & winogrande & 0 & schema \\
Language understanding & bigbench\_language\_identification & 10 & multiple choice \\
\midrule
World knowledge & jeopardy & 10 & language modeling \\
World knowledge & bigbench\_qa\_wikidata & 10 & language modeling \\
World knowledge & arc\_easy & 10 & multiple choice \\
World knowledge & arc\_challenge & 10 & multiple choice \\
\midrule
Commonsense reasoning & copa & 0 & multiple choice \\
Commonsense reasoning & commonsense\_qa & 10 & multiple choice \\
Commonsense reasoning & piqa & 10 & multiple choice \\
Commonsense reasoning & openbook\_qa & 0 & multiple choice \\
\midrule
Symbolic problem solving & bigbench\_dyck\_languages & 10 & language modeling \\
Symbolic problem solving & agi\_eval\_lsat\_ar & 3 & multiple choice \\
Symbolic problem solving & bigbench\_cs\_algorithms & 10 & language modeling \\
Symbolic problem solving & bigbench\_operators & 10 & language modeling \\
Symbolic problem solving & bigbench\_repeat\_copy\_logic & 10 & language modeling \\
\midrule
Reading comprehension & squad & 10 & language modeling \\
Reading comprehension & coqa & 0 & language modeling \\
Reading comprehension & boolq & 10 & multiple choice \\
\bottomrule
\end{tabular}
\caption{
CORE task list. The benchmark contains 22 tasks: 12 multiple-choice tasks, 8 language-modeling tasks, and 2 schema tasks.
}
\label{tab:core_task_list}
\end{table*}

\end{document}